# A Practical Blockchain Framework using Image Hashing for Image Authentication

Cameron C. White, Manoranjan Paul*

*Digital Health and Machine Vision Research Group, School of Computing & Mathematics, Charles Sturt University, Bathurst, NSW 2795, Australia*

ABSTRACT

Blockchain is a relatively new technology that can be seen as a decentralised database. Blockchain systems heavily rely on cryptographic hash functions to store their data, which makes it difficult to tamper with any data stored in the system. A topic that was researched along with blockchain is image authentication. Image authentication focuses on investigating and maintaining the integrity of images. As a blockchain system can be useful for maintaining data integrity, image authentication has the potential to be enhanced by blockchain. There are many techniques that can be used to authenticate images; the technique investigated by this work is image hashing. Image hashing is a technique used to calculate how similar two different images are. This is done by converting the images into hashes and then comparing them using a distance formula. To investigate the topic, an experiment involving a simulated blockchain was created. The blockchain acted as a database for images. This blockchain was made up of devices which contained their own unique image hashing algorithms. The blockchain was tested by creating modified copies of the images contained in the database, and then submitting them to the blockchain to see if it will return the original image. Through this experiment it was discovered that it is plausible to create an image authentication system using blockchain and image hashing. However, the design proposed by this work requires refinement, as it appears to struggle in some situations. This work shows that blockchain can be a suitable approach for authenticating images, particularly via image hashing. Other observations include that using multiple image hash algorithms at the same time can increase performance in some cases, as well as that each type of test done to the blockchain has its own unique pattern to its data.

## 1. Introduction

Cybersecurity is a topic that is increasing in importance as the world continues to develop and rely on technology. An important area in cybersecurity is data integrity. Individuals and industries rely on their data for their daily lives and businesses. A data breach of any kind has the potential to have a severe impact on those who own the data. Therefore, it is important to maintain the security and integrity of data.

A type of data that has risk of being misused and tampered with are images. There are various kinds of images that appear on the internet, such as personal photos and artwork. It is easy for most individuals to be able to tamper with images using photo editing software, such as Photoshop. This allows internet users to upload their own edited images to various websites. The topic of protecting and investigating image integrity is known as image authentication.

Although in most cases these edited images are made for professional work or harmless fun, it is possible for individuals to edit images for malicious purposes. An example would be an individual downloading a copyrighted image, editing it with software, and then uploading the edited image as their own work. [1]

A question that rises from this issue is Is there a method that can be used to detect if an image has been tampered with? Two technologies that can possibly be used to combat this problem are Blockchain and Image Hashing.

Blockchain is a relatively new technology, which became popular in recent times due to it being the technology behind the cryptocurrency bitcoin. A blockchain can be seen as a decentralised database, which means that instead of having a single main database, multiple copies of the same database are used. This creates a network of databases which communicate and work together to collect and use data. A useful security aspect of a blockchain is its ability to maintain data integrity. This is because if one of the databases in a blockchain has its data tampered with, the other databases in the network will notice and will reject the tampered database. [2]

Although blockchain is associated with cryptocurrencies, there has been research into other applications of the technology. Liang and Weller investigate how a blockchain can be used to improve the security of power systems. Whereas in bitcoin the nodes in the blockchain are participating computers, the nodes in the power grid blockchain are electricity meters collecting data on power usage. These meters are placed all over the country and communicate to each other wirelessly, sending encrypted data. The reasoning behind this approach is to decrease the hackability of the meters. [3]

Raje and Vaderia propose a blockchain inspired system for malware detection. In their work they describe a network of devices that acts as a firewall. The nodes in this network each have their own unique malware detection algorithms. Any data that comes into one of the nodes will be broadcasted across the network, allowing the nodes to detect and vote on if the data is malicious. [4]

Image Hashing is a technique that can be used to check if two images are similar to each other. This is done by hashing two images with an image hashing algorithm and then calculating how different the resulting hashes are by using a distance formula.

There are many different image hashing algorithms available to use which hash images in their own unique way, such as Average Hash, Blockmean Hash, and Radial Variance Hash. An issue with image hashing is that there is no single perfect algorithm, as each algorithm has its own strengths and weaknesses. Therefore, it is important that the right algorithm is used for the right job. [5]

It is common for image hashing research to investigate the performance and robustness of image hash algorithms. Zauner investigates and benchmarks multiple image hash algorithms in his work. These algorithms are DCT Based hash, Marr-Hildreth hash, Block Mean hash, and Radial Variance hash. [5]

Drmic and Silic also investigate the performance of image hash algorithms. They compare the robustness of four different image hash algorithms in a variety of scenarios. The algorithms used are average hash, difference hash, P-Hash, and wavelet hash. They concluded that P-Hash was the most robust algorithm out of the four. [6]

Both blockchain and image hashing can be useful for protecting image data. Theoretically, a blockchain can be used to create a database of images that are difficult to tamper with, and image hashing can be used to find if an image has been edited. This can be used to create an image authentication system.

Using these two technologies, a possible solution to the problem could be to create a blockchain database that contains many original, unedited images. This blockchain would have its own image hashing algorithms, which could be used to test the blockchain images against any images given by a user. This would allow users, who suspect that an image they own has been tampered with, to find the most similar image to theirs in the blockchain, allowing them to judge if their image has been edited.

This work aims to investigate this hypothesis by creating an experimental blockchain that implements image hashing. This is done by creating a simulated blockchain in Java, in which the nodes in the blockchain network act as different computers/devices that contain their own unique image hashing algorithm.

When an image is submitted to this blockchain, the devices begin to hash the image, and compare its hash to the images in the blockchain. Using the Hamming distance formula (a formula that is common in image hashing), the devices will find the image that is the least different to the given image. As each device uses their own unique image hashing algorithm, it is possible for the devices to return different images to each other. Due to this issue, the blockchain will favour the device that found the image with the least difference to the given image.

The image hashing algorithms used in this experiment are average hash, phash, blockmean hash, Marr Hildreth hash, and radial variance hash. Each of these algorithms hash images using their own unique method, meaning that their effectiveness in image hashing varies.

The blockchain itself begins with 12 images. Each of these images are unique, meaning that the blockchain cannot accidentally confuse two of the images when no modifications have been applied. 5 of these images will be copied out of the blockchain and then edited with various different methods to be then tested against the blockchain. These editing methods are applying a Gaussian blur, rotating the image, cropping the image, and flipping the image on different axis. The aim behind using these editing methods is to ensure there is a variety of data of the blockchain in different scenarios.

This experiment aims to investigate 3 research questions:

- How can blockchain and image hashing be used together to authenticate images?
- Can the performance of image hash algorithms be improved by using multiple at the same time?
- Is it possible to determine the type of attack done to an image by only looking at image hash data?

## 2. Theory Behind Experiment

### A. Blockchain

In short, a blockchain is a decentralised database. This means that instead of having a single main database, multiple copies of the same database are distributed over a peer to peer network. The intention behind the decentralised approach is to remove the aspect of a central authority in a system. Originating from the paper Bitcoin: A Peer-To-Peer Electronic Cash System [2], a blockchain consists of the following features:
- A peer-to-peer network, consisting of multiple devices.
- A cryptographic hash algorithm (Bitcoin for example uses SHA256).
- A data structure called a *blockchain*.
    - Items in a blockchain are known as blocks. Similar to a linked list, in which every item in the list contains the location of the next, in a blockchain, every block contains the cryptographic hash of the last block in the chain. This means that if a block is tampered with, the blockchain becomes invalid.
    - Each device in the network contains the same copy of the blockchain.
- When new data is being added to the blockchain, the devices on the network must first agree the data is valid before it is added.

### B. Image Authentication

Image authentication is a technique used to verify the integrity of images. There are many software tools available for purchase and for free on the Internet that allows anyone to edit and tamper with images. Although this can be done for professional reasons or for fun, it can also be done for malicious reasons as well. For example, an individual could edit a copyrighted image, upload it to the Internet and then pass it off as their own work. [1]

Hypothetically, an image authentication system could benefit from blockchain. This is because of the aspect of removing trust in a system. An image authentication system could become more reliable and trustworthy by removing the central authority. One particular image authentication technology that could be combined with blockchain is image hashing.

### C. Image Hashing

Image hashing is the technique of comparing images and determining how similar to each other they are. This is done by converting two images into hashes and calculating the difference between the two. Unlike cryptographic hashing, image hashing should not be sensitive to small changes in the input. Image hashing has multiple uses, such as image authentication, image searching, copy detection, etc. [7]

To calculate the difference between two image hashes, a distance formula is used. The formula used by this work is Hamming distance. The Hamming distance formula calculates how many bits are different between two binary strings. [5] This is illustrated in fig 1.

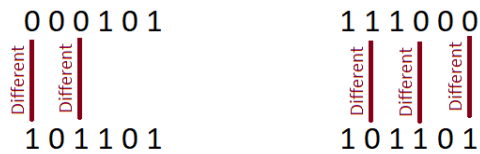

**Fig 1.** A diagram showing the difference between binary strings

Looking at the two binary strings on the left side, it can be noticed that there are only two different bits between the two binary strings. In that example, we would say the two strings have a distance of 2. Whereas in the example on the right, there are three bits different between the two strings, giving it a distance of 3. With this information, it would be stated that the two strings on the left are more similar to each other than the two strings on the right.

*1. Image Hash Algorithms*

There are a variety of different approaches to image hashing, which has caused the creation of many different image hash algorithms. Each image hash algorithm performs differently in different scenarios. For the experiment, image hash algorithms in the OpenCV 3.4.2 library are used (contained in the img_hash module). These algorithms are:
- Average Hash
- P-Hash
- Block Mean Hash
- Marr-Hildreth Hash
- Radial Variance Hash

## 3. Experiment

### A. Approaching the Experiment

When approaching how to design an experiment for combining blockchain and image hashing, the 3 research questions were needed to be considered in the process.

Beginning with the first question, the idea proposed for combining the two technologies was a blockchain system that contains the following features:
- A blockchain network that acts as a database for *original, unedited images*.
- Devices/nodes in the network are assigned their own unique image hash algorithm.
- An image submitted to the blockchain will first be hashed by every node in the network.
    - Each node will compare the hash of the submitted image to every image in the blockchain, finding the most similar image to the submitted one.
    - Due to every algorithm having its own strengths and weaknesses in different scenarios, the node that finds the image with the *least* difference to the submitted image is deemed to be the most correct result. This result is given back to the user.

This design for the blockchain is inspired by Raje, Vaderia, Wilson, & Panigrahi's proposal of a decentralised firewall [4], in which they describe a network where every device contained their own unique malware detection software. This network acts as decentralised firewall. Likewise, in this thesis, a blockchain with nodes containing their own unique image hashing algorithms is proposed.

Using unique image hashing algorithms for each device in the blockchain also relates to the second research question. As each device has a different algorithm, when the blockchain is given an image the nodes all work together on the same input. This means that multiple image hashing algorithms are being used in unison to find the most similar image.

Finally, for the last question, the results gathered from the experiment will be analysed to find any patterns in the data. This is to see if there are any noticeable patterns for different attack scenarios on the images (e.g. Gaussian blur attack).

### B. Experiment Design

For the experiment, there were two possible paths to take. Create an actual blockchain network or develop a simulation of a blockchain. The latter option was taken as it would require less time and resources to create, but still give relevant data for the research questions.

The blockchain simulation was programmed in Java, using the NetBeans IDE (v8.2). The simulation also uses OpenCV 3.4.2 for the image hash algorithms used, as well as to create edited forms of images for the experiment's dataset. Fig 2 displays a UML diagram of the experiment.

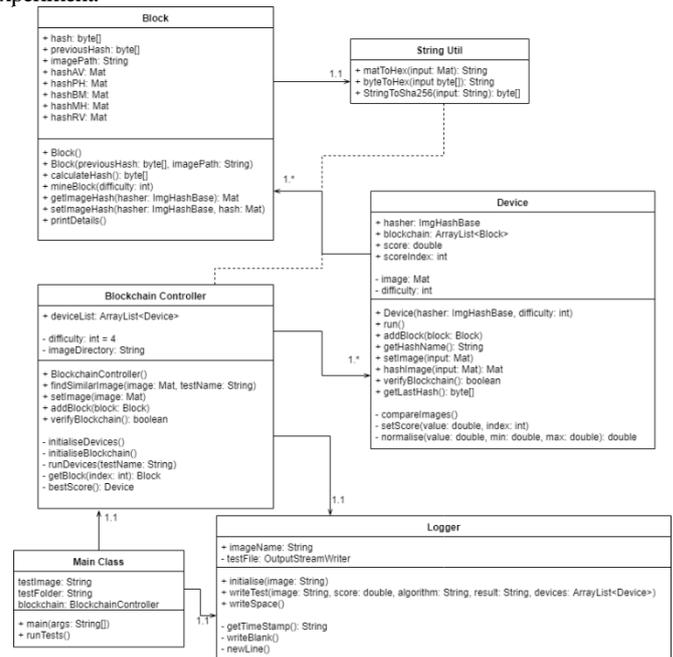

**Fig 2.** A UML diagram of the experiment

The Blockchain Controller is the most vital class of the program, as it acts as the backbone of the simulation. The class functions as a medium for the blockchain to exist in. The blockchain in the program itself consists of multiple devices, which each have an identical copy of the image database and their own unique image hash algorithm. The Logger class is used to write the results of tests into text files for future use.

*1. Blockchain Component*

The blockchain component of the experiment is entirely simulated in Java. It consists of the Blockchain Controller, Device, Block, and String Util classes.

The Blockchain Controller class is what holds the blockchain together. When the program begins, the class starts to generate a simulated blockchain by creating an array of Devices. After that the class will start to add images from the dataset into the blockchain to build it. This process takes some time as it requires each image to be added to a block, which is then mined so that it can be added to the blockchain. Once the blockchain is built, the class can then be used to perform tests on the blockchain. The class will also check if the devices all contain the same blockchain, to ensure it's consistent over devices. These processes are shown in code 1. [8]

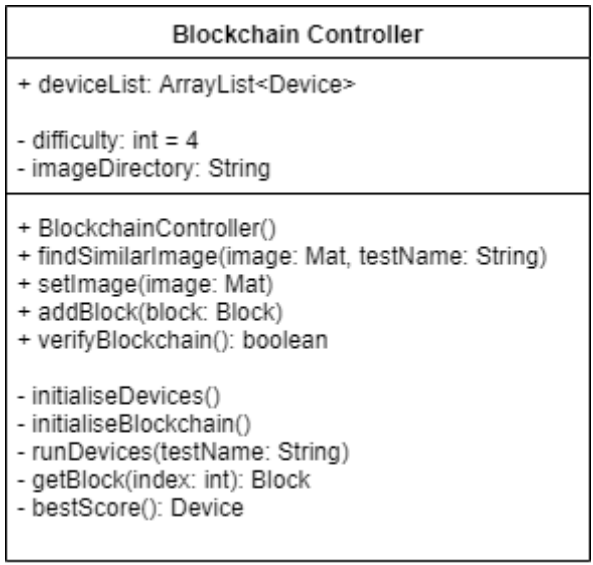

**Fig 3.** The Blockchain Controller Class

```
//Creates a blockchain on start up
  private void initialiseBlockchain() throws IOException{
    //Creates the genesis block (the first block)
    Block genesis = new Block();
    genesis.mineBlock(difficulty);
    addBlock(genesis);

    //List of the images in the given image directory for the
//experiment
    String[] imageList = new File(imageDirectory).list();

    //Adds the other blocks
    for(int i = 0; i < imageList.length; i++){
      String image = imageDirectory + "/" + imageList[i];
      Block block = new Block(
              deviceList.get(0).blockchain.get(i).hash, image);

      Mat blockImage = Imgcodecs.imread(block.imagePath);
      setImage(blockImage);

      //Sets the image hashes for the blocks
      for(int j = 0; j < deviceList.size(); j++){
        Mat hash = deviceList.get(j).hashImage();
        block.setImageHash(
              deviceList.get(j).hasher, hash);
      }

      addBlock(block);
    }
    verifyBlockchain();
  }
```

**Code 1.** How the blockchain is initialised

The Blockchain Controller can be seen as the network of the blockchain. It can be compared to the Internet, as the Internet is a network that is commonly used with blockchains to allow their nodes to communicate to each other. This is illustrated in fig 4.

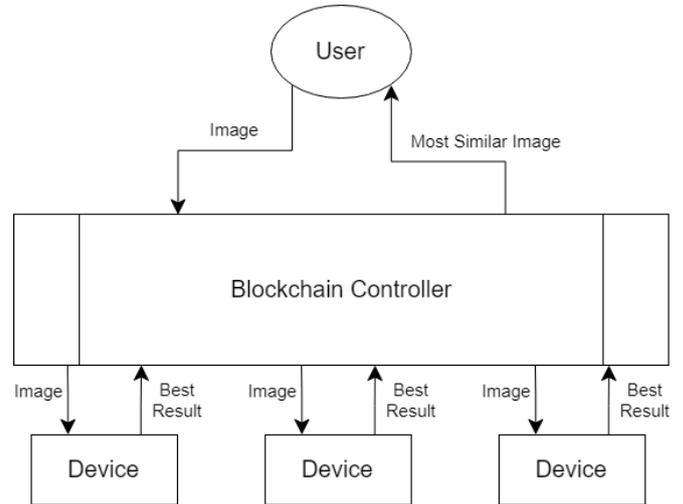

**Fig 4.** A diagram that illustrates the process of searching for the most similar image in the blockchain

The device class acts as a device (or node) on a blockchain network. Each device contains its own unique hashing algorithm and is capable of mining and adding blocks to the blockchain. When an image is given to the blockchain, each device receives a copy of it. This allows the devices to hash the image and (if a test is being performed) find the most similar image in the blockchain. The devices will also check to ensure the blockchain's integrity isn't compromised every time after a block is added. Every device can be run as a separate thread, allowing each device to operate simultaneously.

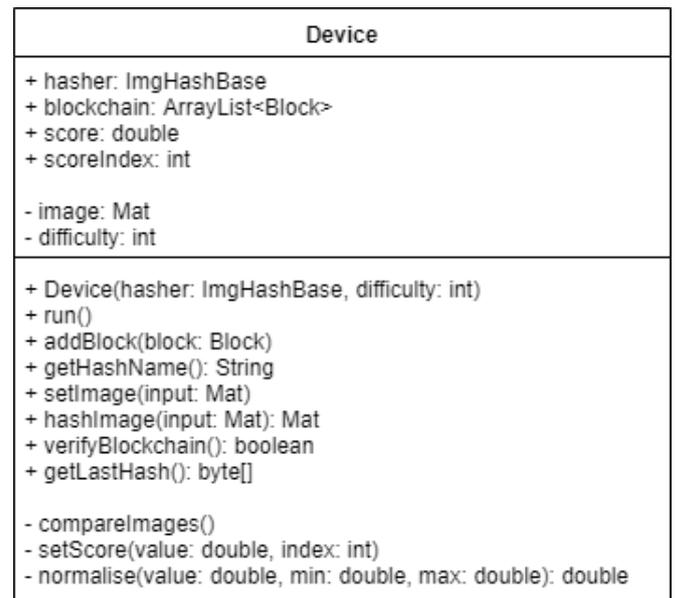

**Fig 5.** The Device Class

The Block class is the most basic component of the blockchain. Each block contains an image, and its respective hashes for every hash algorithm used in the blockchain. Before a block can be added to the blockchain, its cryptographic hash must begin with a particular amount of zeros as according to the current blockchain difficulty level. The difficulty level set for the experiment is 4.

```
Block
────────────────────────────
+ hash: byte[]
+ previousHash: byte[]
+ imagePath: String
+ hashAV: Mat
+ hashPH: Mat
+ hashBM: Mat
+ hashMH: Mat
+ hashRV: Mat
────────────────────────────
+ Block()
+ Block(previousHash: byte[], imagePath: String)
+ calculateHash(): byte[]
+ mineBlock(difficulty: int)
+ getImageHash(hasher: ImgHashBase): Mat
+ setImageHash(hasher: ImgHashBase, hash: Mat)
+ printDetails()
```

**Fig 6.** The Block Class

The purpose of the String Util class is to serve as a tool for converting data between different types as shown in codes 2 and 3. It is most commonly used in the program to create SHA256 cryptographic hashes for blocks, however it can also be used for translating binary data into hex strings, which is useful for debugging and observing the data of the program.

```
String Util
────────────────────────────
+ matToHex(input: Mat): String
+ byteToHex(input byte[]): String
+ StringToSha256(input: String): byte[]
```

**Fig 7.** The String Util Class

```
//Converts a string into a SHA256 hash
    public static byte[] StringToSha256(String input){
      try {
        //Converts the input to SHA-256
        MessageDigest digest =
                MessageDigest.getInstance("SHA-256");

        byte[] hash = digest.digest(input.getBytes("UTF-8"));
        return hash;
      }
      catch (Exception e){
        throw new RuntimeException(e);
      }
    }
```
**Code 2.** Creating a SHA256 hash in Java

```
//Converts data stored in byte arrays (raw binary data) into its
//hex representation
    public static String byteToHex(byte[] input) {
      if(input == null) return "";

      StringBuilder output = new StringBuilder();
      for (int i = 0; i < input.length; i++) {
        String hex = Integer.toHexString(0xff & input[i]);

        if(hex.length() == 1) output.append('0');
        output.append(hex);
      }

      return output.toString();
    }
```

**Code 3.** Converting binary data to a hex string in Java

*2. Image Hashing Component*

The image hashing component of the experiment is performed between the device and block classes. The device class focuses on hashing images using a variety of image hash algorithms, as well as normalising any data resulting from comparing two images. The block class focuses on storing the hash data from the images in the blockchain.

Each device class is assigned its own image hashing algorithm. This is stored in the *hasher* variable. The device uses this variable to hash and compare images (shown in codes 4 and 5), as well as an object used for identification (for example, before normalising data, the class will check which algorithm it uses and will pick a normalisation algorithm based on it). The image hash algorithms used in this experiment are:
- Average hash
- PHash
- Block Mean hash
- Marr-Hildreth hash
- Radial Variance Hash

```
//Returns the hash of the image currently set
  public Mat hashImage(){
    Mat hash = new Mat();
    hasher.compute(image, hash);
    return hash;
  }
```
**Code 4.** Hashing an image in the device class

```
private void compareImages(){
    //Calculates the given image's hash
    Mat imageHash = new Mat();
    hasher.compute(image, imageHash);

    //Scans the blockchain for the most similar image
    double bestScore = 999;
    int mostSimilarIndex = 0;

    for(int i = 1; i < blockchain.size(); i++){
      Mat blockHash = new Mat();
      hasher.compute(
blockchain.get(i).getImageMat(), blockHash);

      //Compares the two images and checks to see if they
      // are the most similar so far
      double result = hasher.compare(imageHash, blockHash);
      if(result < bestScore){
        bestScore = result;
        mostSimilarIndex = i;
      }
    }
    setScore(bestScore, mostSimilarIndex);
  }
```
**Code 5.** Searching the blockchain for the most similar image (condensed)

An important part of the device class is its ability to normalise data. This is to make it easier to perform comparisons between different image hashing algorithms. When two image hashes are compared using the Hamming distance formula, a value is given back. The closer the value is to 0, the more similar the two image hashes are. This value can be between 0 and the maximum possible value from the formula for the two hashes (which is equal to the length of the hash's binary string). Code 6 displays how data is normalised in the experiment, which is based on the following formula:

$$\frac{Result}{Maximum\ Value}$$

```
private double normalise(double value, double max){
    double result;

    result = value / max;

    return result;
}
```
**Code 6.** How data is normalised

In the image hashing component of the experiment, the Block class' purpose is to hold the various hashes of the image it represents. When an image is being converted to a block, it is hashed by every image hashing algorithm existing in the blockchain at the time. Each of these hashes are stored individually in the block. This allows a device to quickly check a particular hash of an image when required without having to rehash the image.

### C. Dataset

For this experiment, a dataset of 12 images are used. These 12 images consist of pictures commonly used in image processing research. They have been sourced from the website Image Processing Place, The dataset used is called "Standard" Test Images. [9]

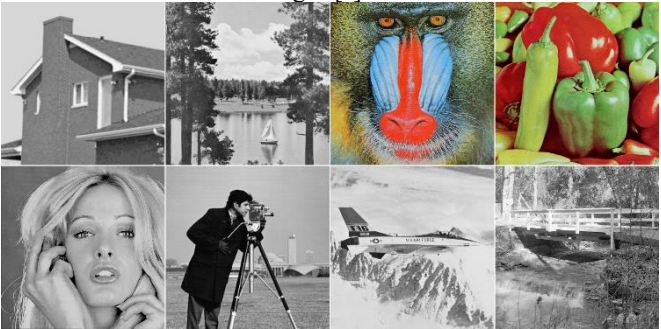

**Fig 8.** 8 of the 12 images contained in the experiment's dataset. The names of the images are as follows (from left to right, top to bottom): House, Lake, Mandrill, Peppers, Woman Blonde, Cameraman, Plane, and Walk Bridge

When the experiment begins, first the blockchain is formed by adding and hashing the dataset. The blockchain will contain every single image in the dataset in individual blocks, along with their hashes for every image hashing algorithm used.

### D. Tests

A test can be performed on the blockchain after it has been formed. A single test is fairly simple; the blockchain is given an image, which it will then hash and compare to the other images in the blockchain. The blockchain will return which image it believes to be the most similar, along with the difference score between the original and the found image. The blockchain will also return the individual data of each device in the system, showing which image it found and its difference score.

To be efficient, the program created for the experiment has been designed to be able to perform tests in bulk. This is done by storing all the images needed to be tested into a folder, which can then be read by the program. This allows the program to test all the images in the folder in one sitting. The results of each test are written to a text file. A sample of the text file is shown in fig 9.

**Fig 9.** A sample of the test data for the Mandrill crop test

To perform tests, a set of test images was created from 6 of the original images in the dataset. These test images are created in bulk, using another program created specifically for this experiment. The program will create "attacked" versions of the image given as input. The attacks performed in this experiment are *Gaussian blur, rotation, cropping,* and *flipping on different axis*. Code 7 gives an example of how one of these attacks (the *Gaussian blur*) are created in the program.

```
public static Mat gaus(Mat image, double value){
    Mat result =
            new Mat(image.rows(), image.cols(), image.type());

    Imgproc.GaussianBlur(image, result, new Size(value,value), 0);

    return result;
}
```
**Code 7.** Sample of creating one of the image attacks; the Gaussian blur

*1. Gaussian Blur*

The Gaussian blur test involves creating a set of 9 images that have had a Gaussian blur filter applied to them. The strength of the blur begins at 5% and increases by 10% for each image created. This is illustrated in fig 10, which shows the image "Peppers" becoming blurrier as it progresses.

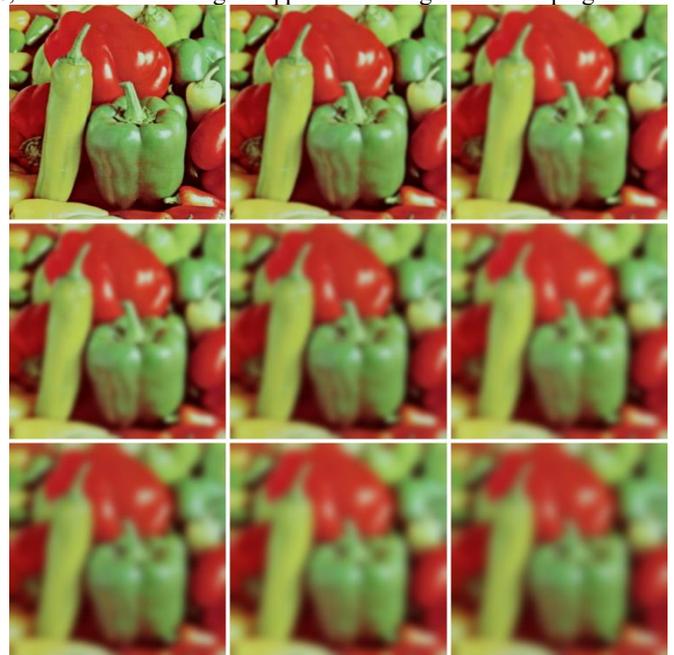

**Fig 10.** An example of the Gaussian blur test images

*2. Rotation*

The rotation test involves creating a set of 9 images that have been rotated by a set amount of degrees. Beginning at 10 degrees, the rotation increases by 10 degrees for each image created. This is illustrated in fig 11, which shows the image "House" being rotated by a different amount in each image.

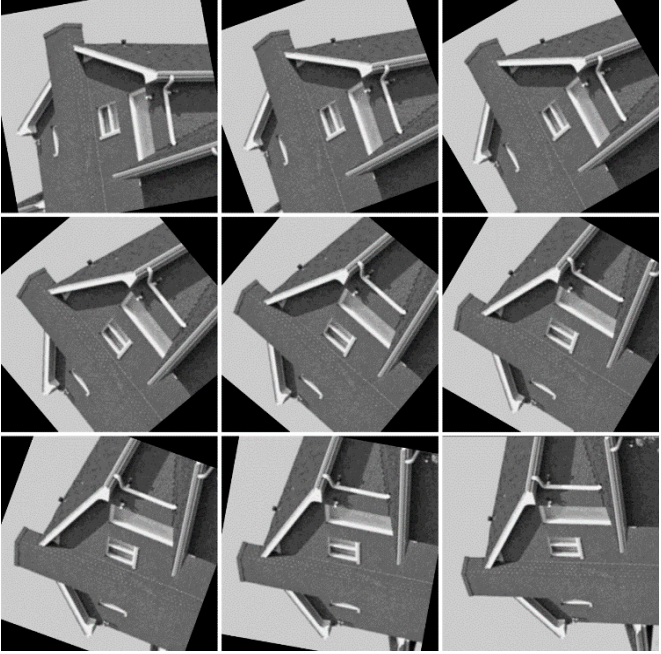

**Fig 11.** An example of the rotation test images

*3. Cropping*

The cropping test involves creating a set of 9 images that have been cropped by a set amount. The amount of the image cropped begins at 10% and increases by 10% for each image created. This is illustrated in fig 12, which shows the image "Mandrill" losing a portion of its content in each image. It should be noted that once an image is cropped, it is not resized to its original resolution (512 x 512), which is not shown in the figure.

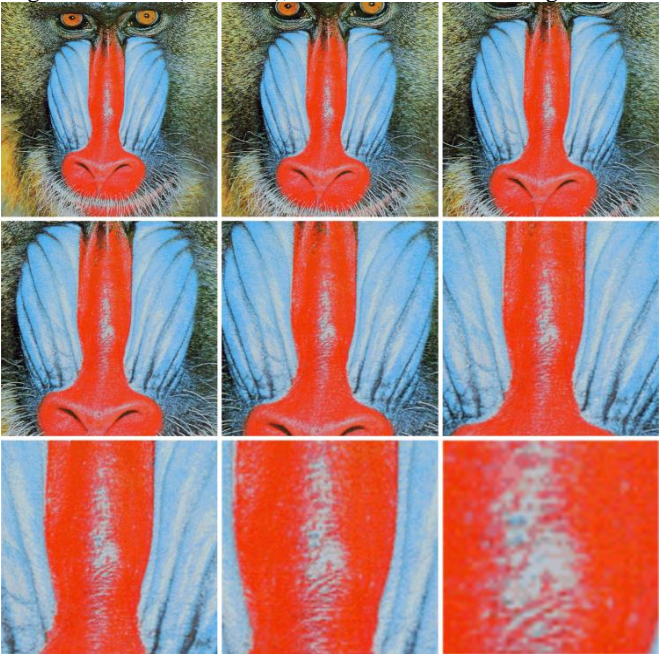

**Fig 12.** An example of the cropping test (images not to scale)

*4. Image Flip*

The image flip test involves creating a set of 3 images that have been flipped on different axis. An image is created for the horizontal, vertical, and both horizontal and vertical axis. This is illustrated in fig 13. The middle image has been flipped horizontally, the right image has been flipped vertically, and the left image has been flipped on both the horizontal and vertical axis.

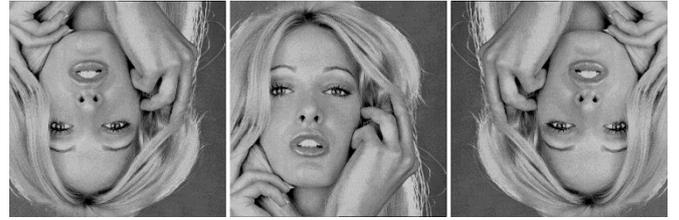

**Fig 13.** An example of the image flip test

### 4. Results

Each image in the test dataset was put through 30 tests, which consist of tampering an image with a particular method and testing it against the blockchain. These tests include:

- 9 Gaussian blur tests; the strength of the blur increases by 10% for each image.
- 9 Rotation tests; the angle of the rotation increases by 10 degrees for each image.
- 9 Crop tests; the amount of content cropped increases by 10% for each image.
- 3 image flip tests; the images created are flipped on the horizontal, the vertical, and both axis.

The results for each test are converted into graphs which are separated into two groups:

- **Blockchain Result:** The blockchain result displays how the blockchain performed overall. It displays if the blockchain returned the correct image, along with the difference score it received. Blue points on the graph indicate that the correct image was returned in the test. Red points indicate that the incorrect image was returned.
- **Individual Algorithm Result:** The individual algorithm result displays how each algorithm performed individually in each test. If an algorithm did not find the correct image in a test, its result is not included.

### A. Gaussian Blur Test

*1. Blockchain Results*

Figs 14 to 18 display how the blockchain performed in the Gaussian blur test. The blockchain performed perfectly in this test as it found the correct image for every single test on every image, as indicated by the blue points on each graph for each image. Notice how every graph is similar, featuring a flat curve. This indicates that a Gaussian blur does not have a severe impact on the blockchain's performance.

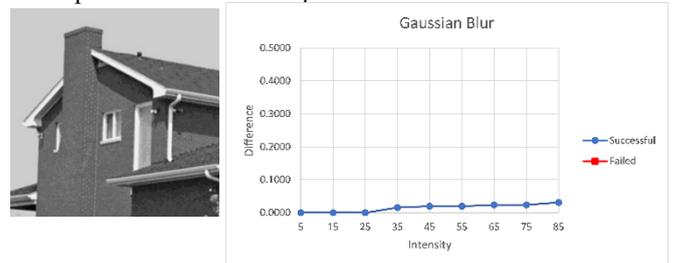

**Fig 14.** House, Gaussian Blur Blockchain Results

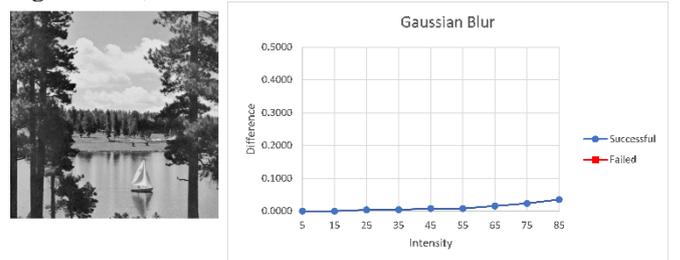

**Fig 15.** Lake, Gaussian Blur Blockchain Results

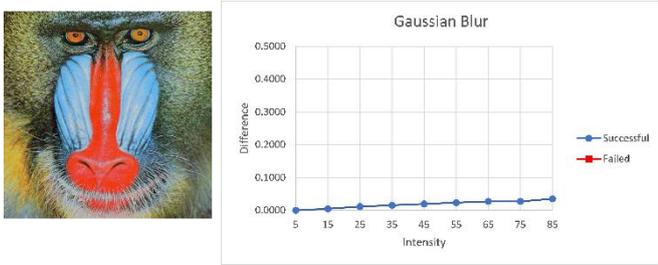
**Fig 16.** Mandrill, Gaussian Blur Blockchain Results

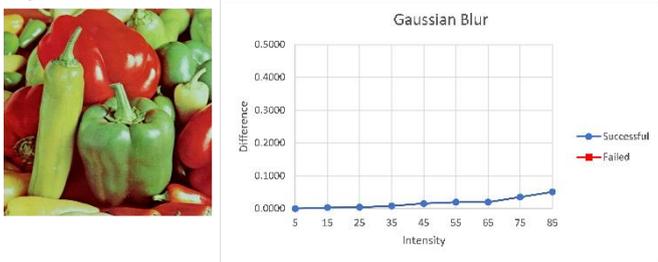
**Fig 17.** Peppers, Gaussian Blur Blockchain Results

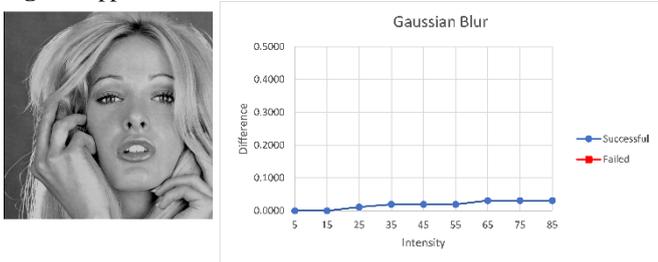
**Fig 18.** Woman Blonde, Gaussian Blur Blockchain Results

*2. Individual Algorithm Results*

Figs 19 to 23 display how the image hash algorithms performed individually in the Gaussian Blur test. All algorithms succeeded in all of the tests (Except radial variance on the 9th Woman Blonde test). However, although the blockchain results gave a mostly flat line graph, the individual results are more varied. Notice that the hashing algorithms perform differently for different images. For example, for the house image in fig 19, average hash has a greater performance than radial variance, but for Mandrill in fig 21, radial variance performs better than average hash.

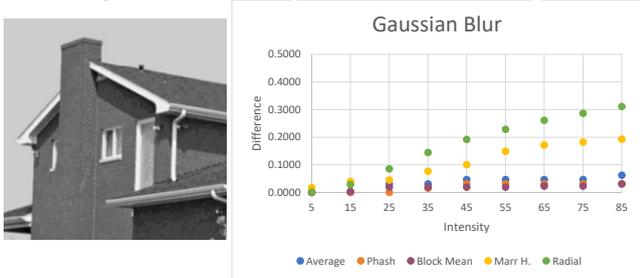
**Fig 19.** House, Gaussian Blur Individual Algorithm Results

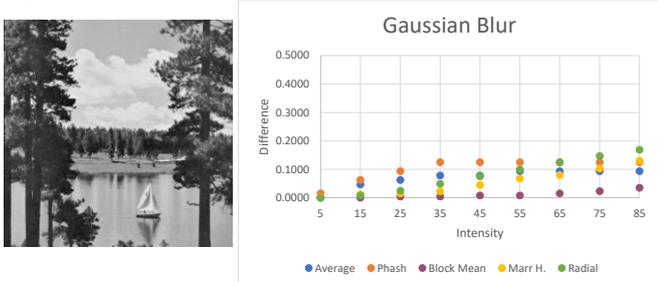
**Fig 20.** Lake, Gaussian Blur Individual Algorithm Results

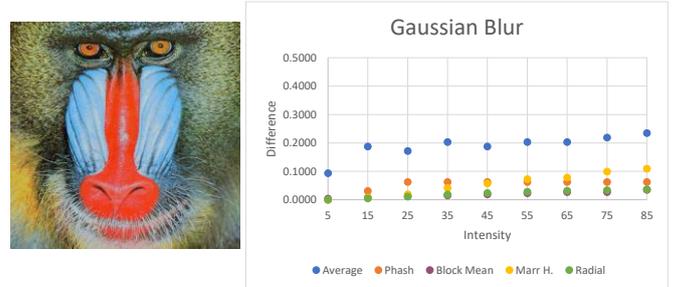
**Fig 21.** Mandrill, Gaussian Blur Individual Algorithm Results

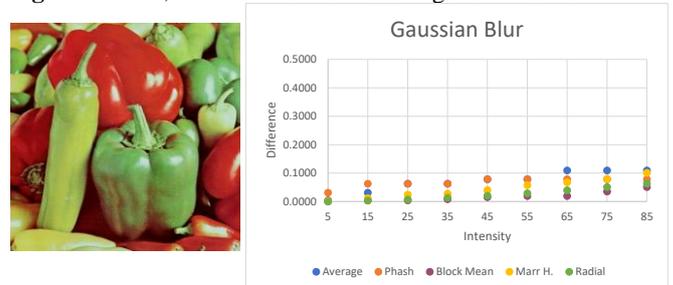
**Fig 22.** Peppers, Gaussian Blur Individual Algorithm Results

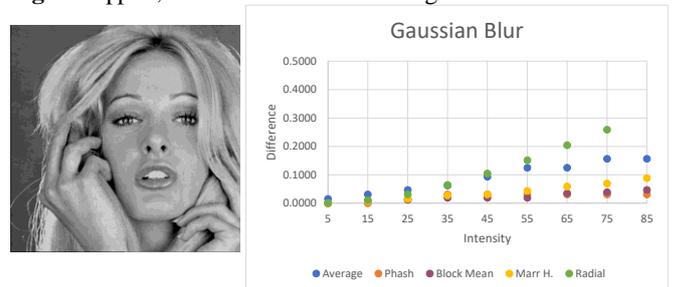
**Fig 23.** Woman Blonde, Gaussian Blur Individual Algorithm Results

**B. Rotation Test**

*1. Blockchain Results*

Figs 24 to 28 display how the blockchain performed in the Rotation test. The blockchain performed poorly in this test as it retuned incorrect images for most of the tests (this is indicated by the red squares on the graphs). However, all images passed the first rotation test, except for peppers in fig 27. The difference scores for each test are seemingly random and shows no pattern.

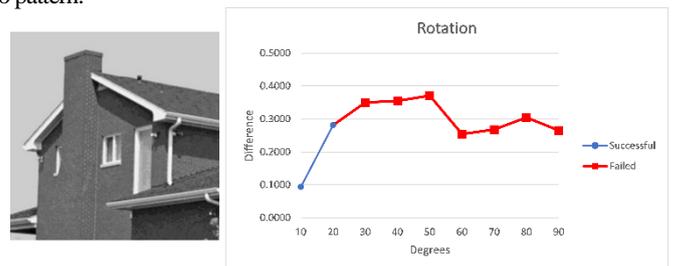
**Fig 24.** House, Rotation Test Blockchain Results

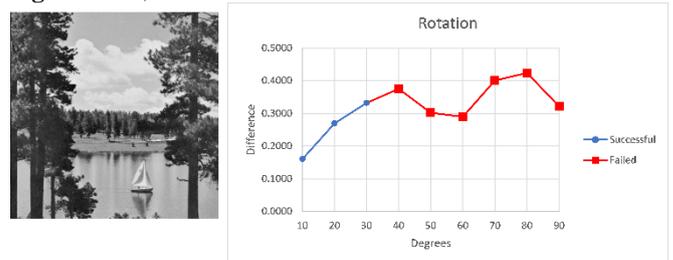
**Fig 25.** Lake, Rotation Test Blockchain Results

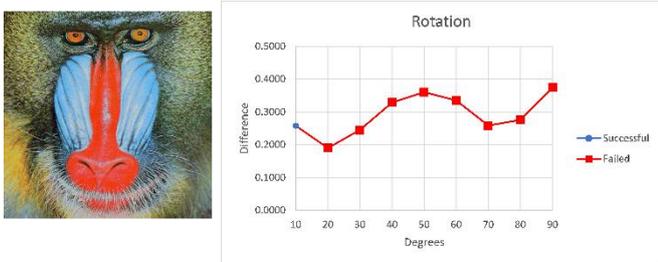
**Fig 26.** Mandrill, Rotation Test Blockchain Results

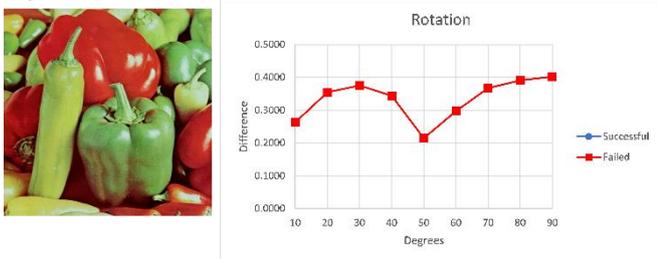
**Fig 27.** Peppers, Rotation Test Blockchain Results

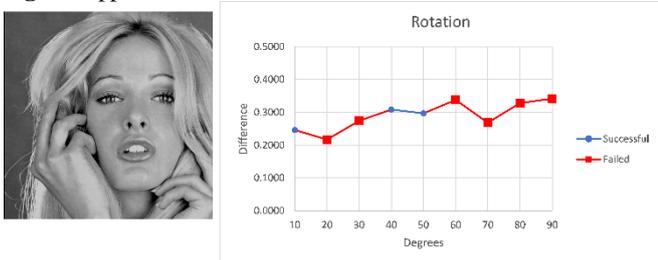
**Fig 28.** Woman Blonde, Rotation Test Blockchain Results

*2. Individual Algorithm Results*

Figs 29 to 33 display how the image hash algorithms performed individually in the Rotation Test. Although the blockchain struggled greatly with the rotation test, the algorithms individually performed moderately in some cases, such as for house (fig 29) and blonde woman (fig 33). The exception however is peppers (fig 32), in which both the blockchain and the algorithms struggled.

It must be noted that if an algorithm gave an incorrect result in a test, its data would not be added to the graph.

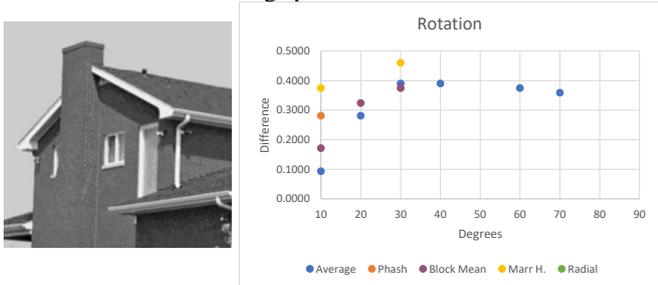
**Fig 29.** House, Rotation Test Individual Algorithm Results

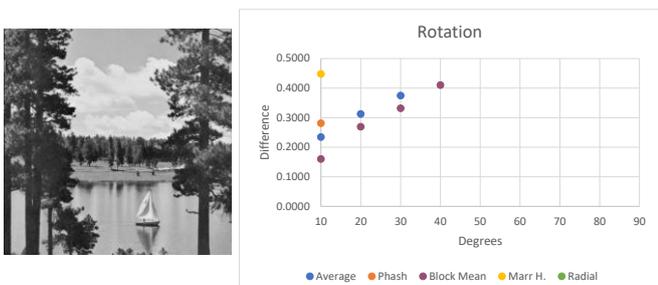
**Fig 30.** Lake, Rotation Test Individual Algorithm Results

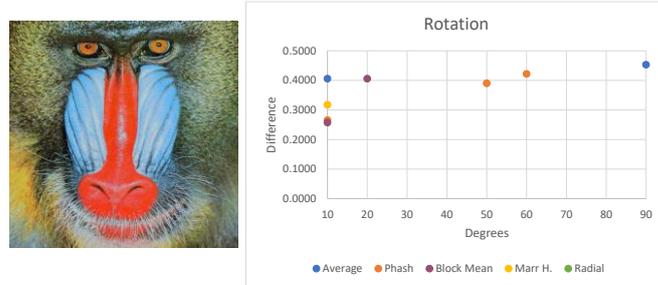
**Fig 31.** Mandrill, Rotation Test Individual Algorithm Results

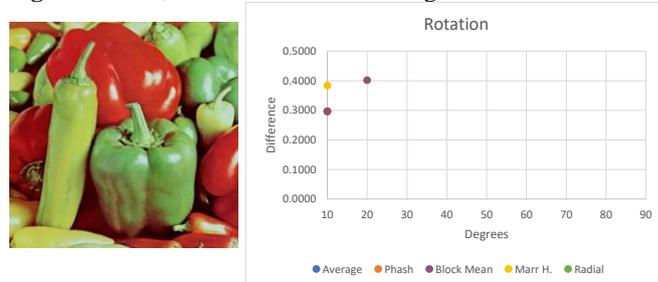
**Fig 32.** Peppers, Rotation Test Individual Algorithm Results

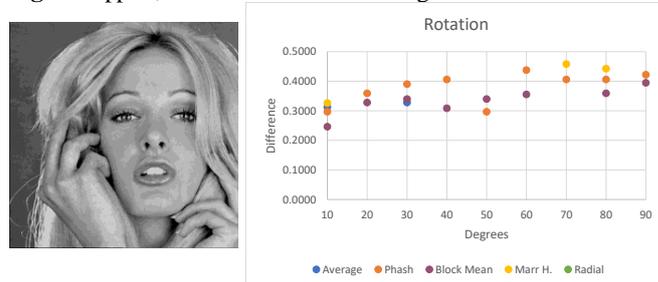
**Fig 33.** Woman Blonde, Rotation Test Individual Algorithm Results

## C. Crop Test

*1. Blockchain Results*

Figs 34 to 38 display how the blockchain performed in the Crop test. Like the rotation test, the crop test did return incorrect images at some points. However, the blockchain did perform more reliably and predictably in this test. All images give a similar, gradual curve in their data.

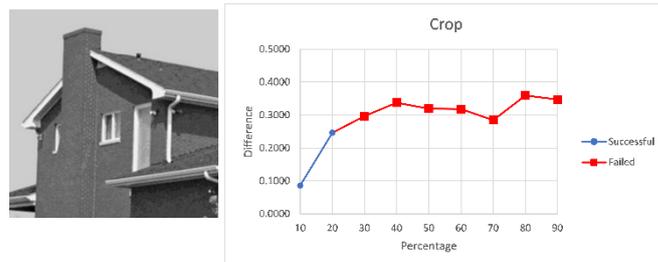
**Fig 34.** House, Crop Test Blockchain Results

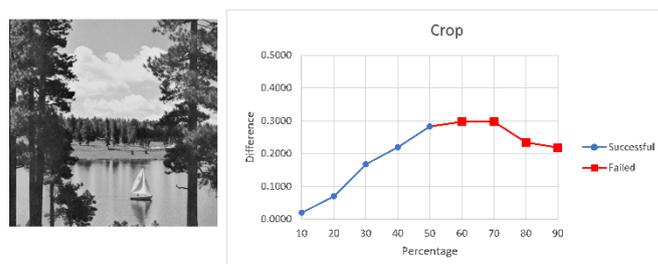
**Fig 35.** Lake, Crop Test Blockchain Results

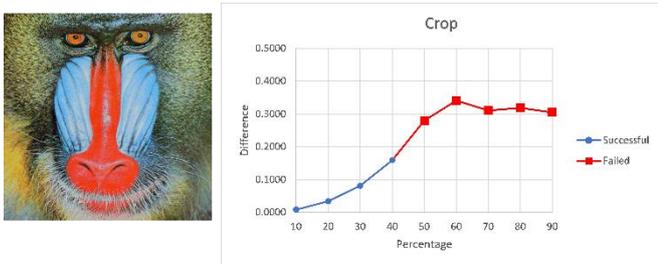
**Fig 36.** Mandrill, Crop Test Blockchain Results

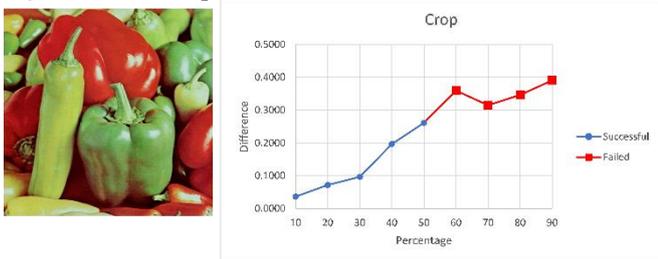
**Fig 37.** Peppers, Crop Test Blockchain Results

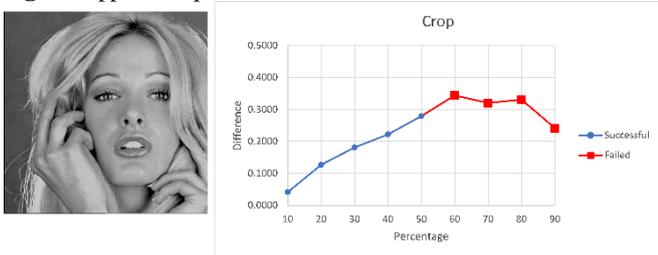
**Fig 38.** Woman Blonde, Crop Test Blockchain Results

*2. Individual Algorithm Results*

Figs 39 to 43 display how the image hash algorithms performed individually in the Crop Test. The main observation for these results is that radial variance is the dominant algorithm in this test and shows a similar curve in each graph to its respective blockchain counterpart.

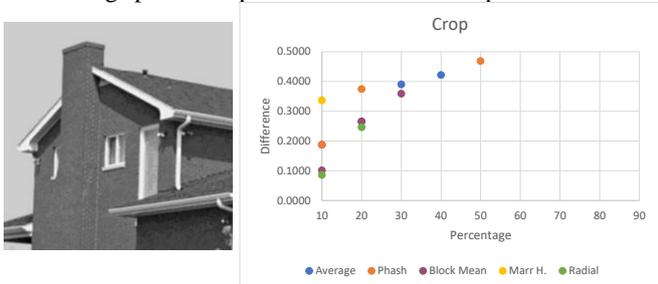
**Fig 39.** House, Crop Test Individual Algorithm Results

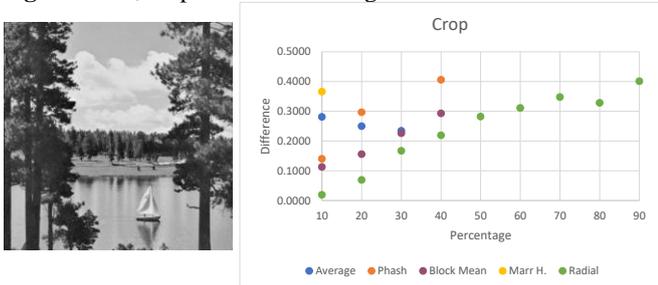
**Fig 40.** Lake, Crop Test Individual Algorithm Results

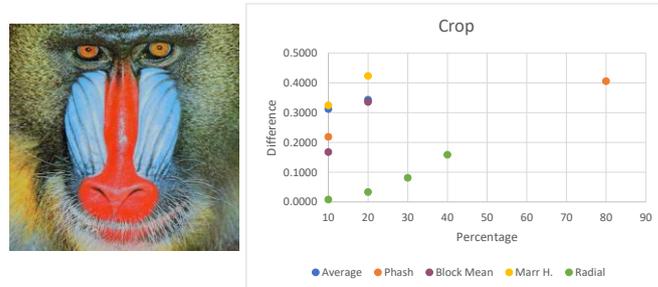
**Fig 41.** Mandrill, Crop Test Individual Algorithm Results

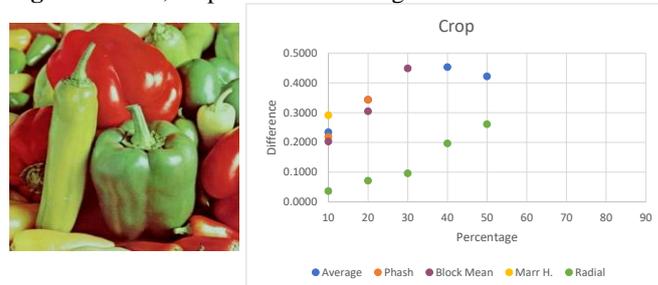
**Fig 42.** Peppers, Crop Test Individual Algorithm Results

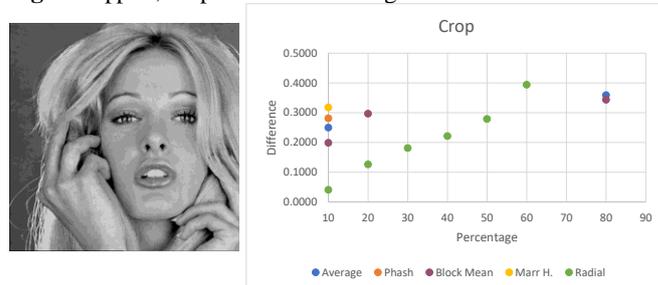
**Fig 43.** Woman Blonde, Crop Test Individual Algorithm Results

### D. Image Flip Test

*1. Blockchain Results*

Figs 44 to 48 display how the blockchain performed in the Image Flip test. The blockchain performed well in this test but gave an incorrect result in some cases such as house (fig 44), peppers (fig 47), and blonde woman (fig 48). The two main observations for this test are that flipping the image either horizontally or vertically yield similar results in most cases, and that flipping the image in both directions gives results close to 0.

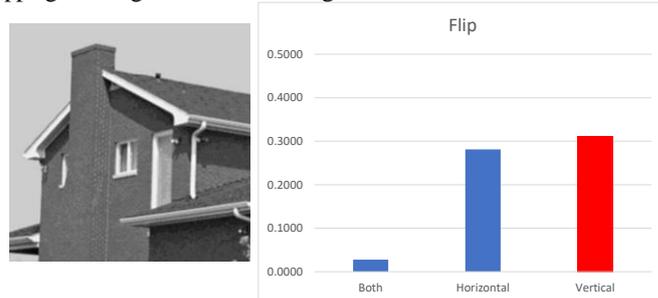
**Fig 44.** House, Image Flip Test Blockchain Results

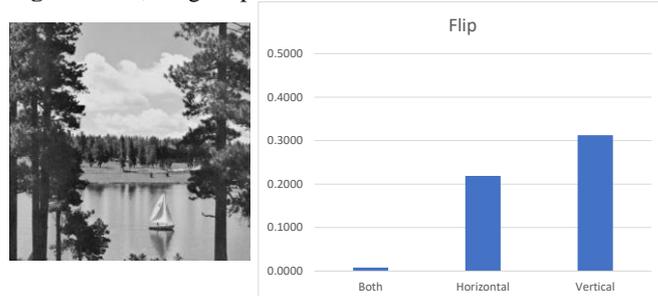
**Fig 45.** Lake, Image Flip Test Blockchain Results

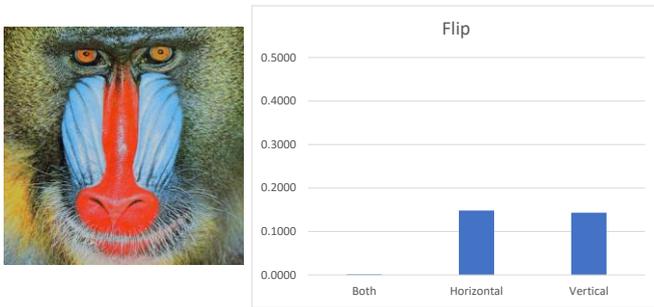
**Fig 46.** Mandrill, Image Flip Test Blockchain Results

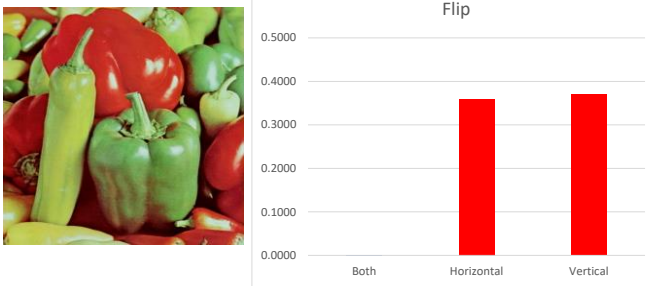
**Fig 47.** Peppers, Image Flip Test Blockchain Results

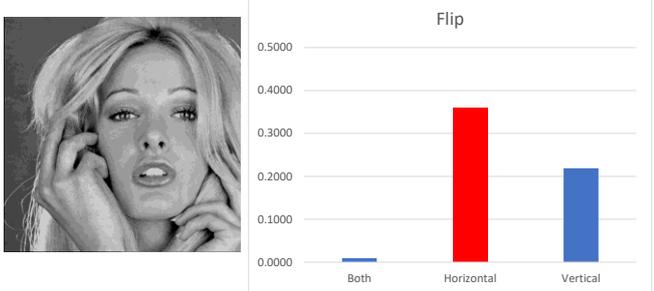
**Fig 48.** Woman Blonde, Image Flip Test Blockchain Results

*2. Individual Algorithm Results*

Fig 49 displays how the image hash algorithms performed individually in the image flip test. If an algorithm returned the incorrect result, its data would not be added to the table. The main observation is that it appears to be random if an algorithm succeeds or not, although radial variance appears to be the most consistent.

## 5. Analysis and Discussion

### A. Gaussian Blur

The Gaussian blur was the test that the blockchain handled the best. The blockchain found the correct image every single time without fail, and the difference score received for each test was always below 0.1. This means that the blockchain considers the given and found images to be very similar to each other.

Individually, the hashing algorithms performed well as none of them found the incorrect image (an exception being radial variance on woman blonde), although some algorithms would perform better than others in different cases. The performance of each algorithm appears to vary between image to image. However, *block mean hash* appears to be the most consistent of the algorithms, as its difference score rarely rises above 0.05 in the tests, whereas the other algorithms (such as radial variance) excel with some images, but struggle with others.

By looking at the graphs alone a Gaussian blur test has a very obvious pattern, as its data forms a slight, almost flat, curve for every image.

### B. Rotation

The rotation attack was the test that the blockchain struggled the most with. The blockchain would return the incorrect image in most cases.

Although the blockchain itself struggled with the rotation attack. In some cases, the hashing algorithms could actually find the correct image, albeit with high difference scores in most cases. The image that the algorithms performed the best on was *Blonde Woman* (fig 33), with *pHash* and *block mean hash* in particular performing well. *Peppers* (fig 32) was the image the algorithms struggled with the most, with few of the tests returning positive results.

The data from the rotation test appears to be seemingly random (although negative results are fairly consistent). This make the rotation test easy to identify by just looking at the data alone.

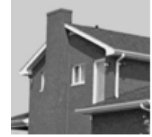
House
| Direction | Average | Phash | Block Mean | Marr H. | Radial |
|---|---|---|---|---|---|
| Both | | | | | 0.0278 |
| Horizontal | 0.2813 | | | | |
| Vertical | | | | | |

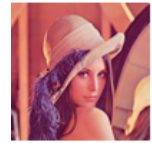
Lena
| Direction | Average | Phash | Block Mean | Marr H. | Radial |
|---|---|---|---|---|---|
| Both | | | | | 0.0031 |
| Horizontal | | | | | 0.3432 |
| Vertical | 0.3438 | | 0.3672 | 0.4306 | 0.3431 |

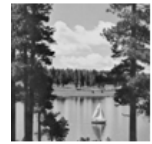
Lake
| Direction | Average | Phash | Block Mean | Marr H. | Radial |
|---|---|---|---|---|---|
| Both | | | 0.3750 | | |
| Horizontal | 0.2188 | | 0.2344 | 0.3438 | |
| Vertical | | | 0.3125 | | |

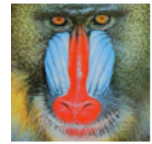
Mandrill
| Direction | Average | Phash | Block Mean | Marr H. | Radial |
|---|---|---|---|---|---|
| Both | | | | | 0.0013 |
| Horizontal | | | 0.2578 | 0.2431 | 0.1483 |
| Vertical | 0.4063 | | | 0.4479 | 0.1432 |

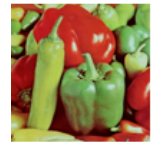
Peppers
| Direction | Average | Phash | Block Mean | Marr H. | Radial |
|---|---|---|---|---|---|
| Both | 0.2813 | | 0.4297 | 0.4514 | 0.0014 |
| Horizontal | | | 0.4297 | | |
| Vertical | | | | | |

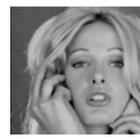
Blonde Woman
| Direction | Average | Phash | Block Mean | Marr H. | Radial |
|---|---|---|---|---|---|
| Both | | | | | 0.0093 |
| Horizontal | | 0.3906 | | 0.4531 | |
| Vertical | 0.3125 | 0.2813 | 0.2188 | 0.3976 | |

**Fig 49.** Image Flip Individual Algorithm Results (All Images)

### C. Crop

In most cases, the blockchain could handle the crop test fairly well. All of the images can be successfully matched with the correct image for the beginning of each test, and that success can be maintained for at least the first 4 tests for each image (except for *house* (fig 34), which the blockchain oddly struggled with).

Looking at the hashing algorithms individually however gives a somewhat more mixed result. The best performing and most consistent algorithm appears to be radial variance, making it the more reliable algorithm in this scenario. In most of the cases, the algorithms lose their effectiveness when the image has been cropped by 50% (varying from algorithm to algorithm).

The data from the crop test forms a gradual curve in a line graph, until the blockchain begins to give false results, in which the difference score begins to even out. It is reasonable to be able to identify a crop test by data by a gradual curve that begins to even out at the end.

### D. Image Flip

The image flip test gave interesting results. In a few cases, the *vertical,* and the *horizontal* axis flip tests gave very similar difference scores. What was more interesting however, was that flipping the image on both axis gives difference scores that are very close to 0 in all cases.

However, although the blockchain performed well in most cases, many of the image hash algorithms struggled with the image flip test. The most consistent of the algorithms was Radial variance, though it still did struggle in some cases, like the *lake* image (fig 45).

The pattern of the data is easily identifiable by noticing that the vertical, and the horizontal tests give similar scores, and that flipping images on both axis give scores close to 0.

### E. Overview

Overall, the blockchain has a moderate performance across the different tests. It is clear that the blockchain excels in some scenarios (such as Gaussian blur) and struggles in others (such as a rotation attack). This is similar to how image hashing algorithms have their own strengths and weaknesses, which is also demonstrated by the experiment by looking at the individual performance of the image hash algorithms.

When deciding which image to use for each test, the blockchain will pick the algorithm (device) which gives the lowest difference score. Through the results it is evident that this is not a strong strategy to use, as even if some algorithms find the correct the image, the blockchain may favour an incorrect algorithm that gives a lower score.

An example of this is the crop test using the lake image. Looking at its overall results, the radial variance algorithm (fig 39) could find the correct image every single time. However, the blockchain (fig 34) itself only returned the correct result 5 times out of the 9 tests.

By looking at the data for each test, it is obvious that each scenario has its own unique pattern for its data. This could be useful for identifying how image has been tampered with.

## 6. Conclusion

Overall the experiment was a success. The program that was created works as a proof of concept for creating an image authentication framework using blockchain and image hashing. Although the system developed is not perfect and struggles in some situations, it proves that combining blockchain and image hashing is possible and lays the foundation for future research.

### A. Answering Research Questions

There were three research questions for this experiment, which have all received some form of an answer to. These were:
1. How can blockchain and image hashing be used together to authenticate images?
2. Can the performance of image hash algorithms be improved by using multiple at the same time?
3. Is it possible to determine the type of attack done to an image by only looking at image hash data?

For the first question, a framework was proposed in which a blockchain can authenticate images using image hashing. This is done by having a blockchain act as a database for images, which is made up of nodes that contain their own image hashing algorithm. A user can use this blockchain to authenticate images by submitting their own images to it. The nodes in the blockchain will hash the image and compare it to the image database and the best result from the nodes is then given to the user.

As shown by the experiment, theo method proposed is functional. However, it is not perfect, as it is shown through the results that this system is not robust against every situation (e.g. rotation and cropping attack). It is evident that picking the node that calculates the lowest difference score is not the best solution to picking the result.

For the second question, the experiment indicates that image hashing algorithms can be enhanced by using them in unison. This is evident from the results as it is clear that some image hashing algorithms perform better in some scenarios than others. For example, *radial variance* is shown in the results to be the best performing image hash for the cropping test. However, for the Gaussian blur test, *block mean hash* was the most reliable. If either of these algorithms were removed from the experiment, or if only one algorithm was used, the blockchain would have performed considerably worse.

Finally, for the last question, from the results it is clear that each test had its own unique pattern for its data. Gaussian blur gave a near flat curve, rotating gave seemingly random results, cropping gave a gradual curve, and flipping the image in different directions gave particularly distinct results. Therefore, by analysing the patterns in the data, it is possible to identify how an image was tampered.

### B. Paths for Future Research

One of the important things learned from this experiment is that there is much more potential research in this topic. There are many possible improvements that can be researched and added to the proposed framework, such as:

- **What is the best strategy for picking the most reliable result from the blockchain?**
  It is clear from the experiment that picking the node that gives the lowest difference score is not the most reliable method for determining the result of a test. There needs to be research done to find out how a blockchain can pick the most reliable algorithm/node for the result.

- **Are there any other image authentication techniques that can be used with blockchain?**
  Image hashing is not the only image authentication technique that can be used with blockchain. There are many other methods that could be experimented on, such as watermarking. [1]

- **What other image hashing algorithms can be used for this framework?**
  For this experiment, only five image hashing algorithms were used. There are many other image hashing algorithms that could be implemented into the system, such as colour moment [10] or ring partition. [7]

- **What other tests can be performed on the blockchain?**
  Due to time constraints, only four different types of tests were performed on the blockchain (Gaussian blur, rotation, cropping, and image flipping). There are many more methods that can be done to attack an image (such as JPEG compression, salt and pepper attack, etc.).